\documentclass[runningheads]{llncs}
\usepackage{graphicx}
\usepackage{subcaption}
\usepackage{booktabs}
\usepackage[ruled,vlined]{algorithm2e}
\usepackage{amsfonts}
\def\header{\vspace{1mm} \noindent}
\usepackage[misc]{ifsym}
\begin{document}
\title{LSCALE: Latent Space Clustering-Based Active Learning for Node Classification}
\titlerunning{Latent Space Clustering-Based Active Learning for Node Classification}
\toctitle{Latent Space Clustering-Based Active Learning for Node Classification}
%
\author{Juncheng Liu (\Letter) \and Yiwei Wang \and Bryan Hooi \and Renchi Yang \and Xiaokui Xiao}
\tocauthor{Juncheng~Liu, Yiwei~Wang, Bryan~Hooi, Renchi~Yang, and Xiaokui~Xiao}
\authorrunning{J. Liu et al.}
\institute{School of Computing, National University of Singapore, Singapore\\
\email{\{juncheng, y-wang, bhooi\}@comp.nus.edu.sg} \\
\email{\{renchi, xkxiao\}@nus.edu.sg}
}
%
\maketitle              
\begin{abstract}
Node classification on graphs is an important task in 
many practical domains. It usually requires labels for training, which can be difficult or expensive to obtain in practice. 
Given a budget for labelling, active learning aims to improve performance by carefully choosing which nodes to label. 
Previous graph active learning methods learn representations using labelled nodes and select some unlabelled nodes for label acquisition. However, they do not fully utilize the representation power present in unlabelled nodes. 
We argue that the representation power in unlabelled nodes can be useful for active learning and for further improving performance of active learning for node classification. 
In this paper, we propose a latent space clustering-based active learning framework for node classification (LSCALE), where we fully utilize the representation power in both labelled and unlabelled nodes.
Specifically, to select nodes for labelling, our framework uses the K-Medoids clustering algorithm on a latent space based on a dynamic combination of both unsupervised features and supervised features. In addition, we design an incremental clustering module to avoid redundancy between nodes selected at different steps. 
Extensive experiments on five datasets show that our proposed framework LSCALE consistently and significantly outperforms the state-of-the-art approaches by a large margin. 

\end{abstract}

\section{Introduction} 
Node classification on graphs has attracted much attention in the graph representation learning area. 
Numerous graph learning methods \cite{semi_GCN,hamilton2017inductive_Graphsage,SGC,chen2018fastgcn} have been proposed for node classification with impressive performance, especially on the semi-supervised setting, where labels are required for the classification task. 

In reality, labels
are often difficult and expensive to collect.
To mitigate this issue, active learning aims to select the most informative data points which can lead to better classification performance using the same amount of labelled data. 
Graph neural networks (GNNs) have been used 
for some applications such as disease prediction and drug discovery \cite{parisot2018disease,pmlr-v70-gilmer17a_Quantum_Chemistry}, in which labels often have to be obtained through costly means such as chemical assays.
Thus, these applications motivate research into active learning with GNNs.

In this work, we focus on active learning for node classification on attributed graphs. 
Recently, a few GNN-based active learning methods \cite{active_graph_embedding,ANRMAB,wu2019active_featprop,ICML-regol20-active,transferable-active-learning} have been proposed for attributed graphs.
However, their performance is still less than satisfactory in terms of node classification. 
These approaches do not fully utilize the useful representation power in unlabelled nodes and only use unlabelled nodes for label acquisition. 
For example, AGE \cite{active_graph_embedding} and ANRMAB \cite{ANRMAB} select corresponding informative nodes to label based on the hidden representations of graph convolutional networks (GCNs) and graph structures. 
These hidden representations can be updated only based on the labelled data. 
On the other hand, FeatProp \cite{wu2019active_featprop} is a clustering-based algorithm which uses propagated node attributes to select nodes to label. 
However, these propagated node attributes are generated in a fixed manner based on the graph structure and node attributes, and are not learnable.
In summary, existing approaches do not fully utilize the information present in unlabelled nodes. 
To utilize the information in unlabelled nodes, a straightforward method is to use features extracted from a trained unsupervised model for choosing which nodes to select. 
For example, FeatProp can conduct clustering based on unsupervised features for selecting nodes. However, as shown in our experimental results, it still cannot effectively utilize the information in unlabelled nodes for active learning. 

Motivated by the limitations above, 
we propose an effective \textbf{L}atent \textbf{S}pace \textbf{C}lustering-based \textbf{A}ctive \textbf{LE}arning framework (hereafter \textbf{LSCALE}). 
In this framework, we conduct clustering-based active learning on a designed latent space for node classification. Our desired \textit{active learning latent space} should have two key properties: 1) \textit{low label requirements:} it should utilize the representation power from all nodes, not just labelled nodes, thereby obtaining accurate representations even when very few labelled nodes are available; 2) \textit{informative distances:} in this latent space, intra-class nodes should be closer together, while inter-class nodes should be further apart. This can facilitate clustering-based active selection approaches, which rely on these distances to output a diverse set of query points.

To achieve these, our approach incorporates an unsupervised model (e.g., DGI \cite{DGI}) on all nodes to generate unsupervised features, which utilizes the information in unlabelled nodes, satisfying the first desired property. 
In addition, we design a distance-based classifier to classify nodes using the representations from our latent space. 
This ensures that distances in our latent space are informative for active learning selection, satisfying our second desired property. 
To select nodes for querying labels, we leverage the K-Medoids clustering algorithm in our latent space to obtain cluster centers, which are the queried nodes. 
As more labelled data are received, the distances between different nodes in the latent space change based on a dynamic combination of unsupervised learning features and learnable supervised representations. 


Furthermore, we propose an effective incremental clustering strategy for clustering-based active learning to prevent redundancy during node selection. Existing clustering-based active learning methods like \cite{core-set,wu2019active_featprop} only select nodes in multiple rounds with a myopic approach. 
More specifically, in each round, they apply clustering over all unlabelled nodes and select center nodes for labelling. 
However, the cluster centers tend to be near the ones obtained in the previous rounds.
Therefore, the clustering can select redundant nodes and does not provide much new information in the later rounds. 
In contrast, our incremental clustering is designed to be aware of the selected nodes in the previous rounds and ensure newly selected nodes are more informative. 
Our contributions are summarized as follows:
\vspace{-\topsep}
\begin{itemize}
\item We propose a latent space clustering-based active learning framework (LSCALE) for node classification on attributed graphs. 
LSCALE contains a latent space with two key properties designed for active learning purposes: 1) \emph{low label requirements}, 2) \emph{informative distances}. 
\item We design an incremental clustering strategy to ensure that newly selected nodes are not redundant with previous nodes, which further improves the performance.
\item We conduct comprehensive experiments on three public citation datasets and two co-authorship datasets. The results show that our method provides a consistent and significant performance improvement compared to the state-of-the-art active learning methods for node classification on attributed graphs. 
\end{itemize}
\begin{figure}[t]
	\centering
	\includegraphics[width=0.9\columnwidth]{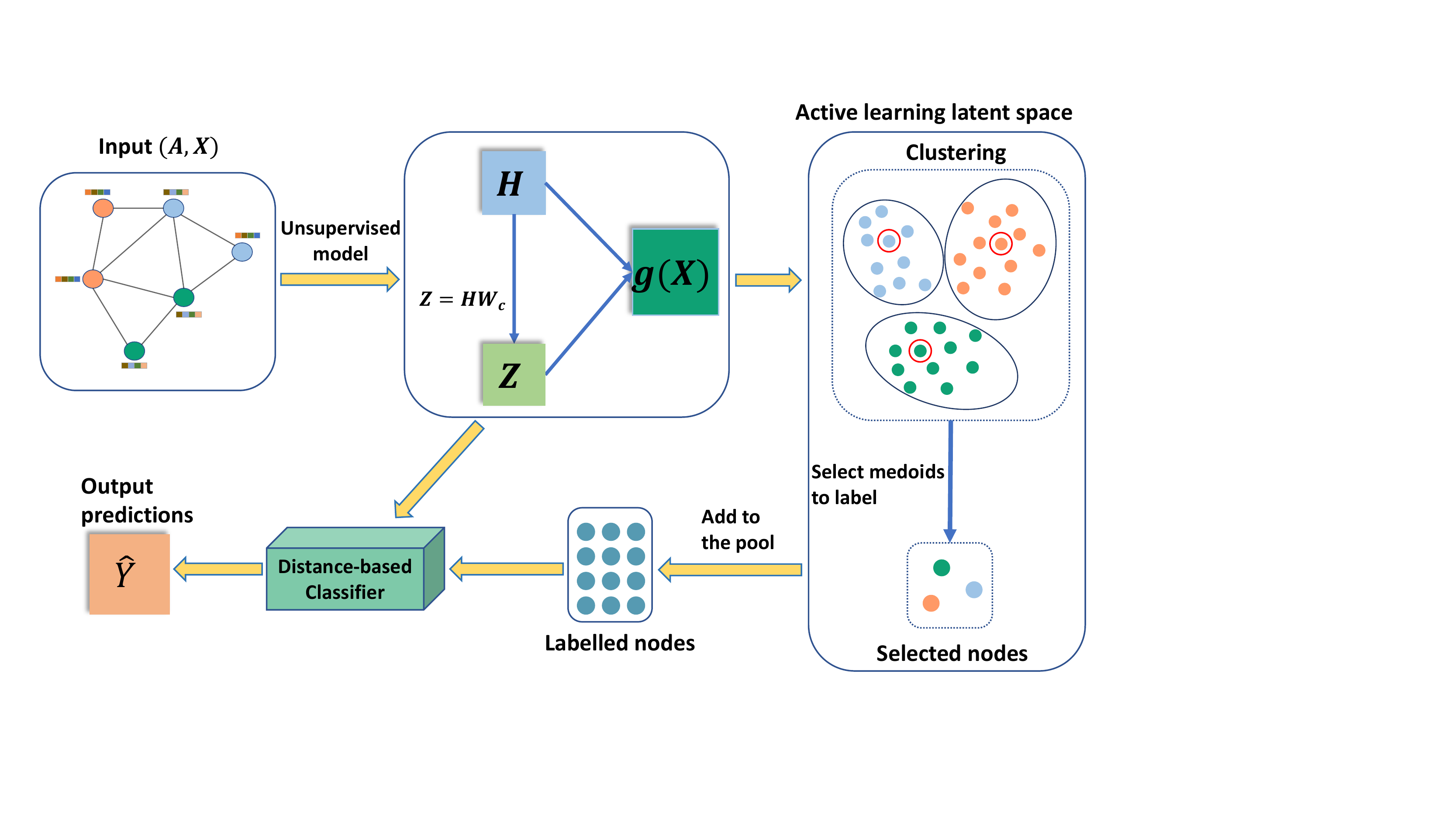}
	\caption{Illustration of the proposed framework LSCALE.}
	\label{fig:framework}
\end{figure}
\section{Problem Definition}
In this section, we present the formal problem definition of active learning for node classification. 
Let $G = (V,E)$ be a graph with node set $V$ and edge set $E$, where $|V|=n$ and $|E|=m$.
$\mathbf{X} \in \mathbb{R}^{n\times d}$ and $\mathbf{Y}$ represent the input node attribute matrix and label matrix of graph $G$, respectively. In particular, each node $v\in V$ is associated with a length-$d$ attribute vector $\mathbf{x}_v$ and a one-hot label vector $\mathbf{y}_v$.
Given a graph $G$ and its associated attribute matrix $\mathbf{X}$, node classification aims to find a model $\mathcal{M}$ which predicts the labels for each node in $G$ such that the loss function $\mathcal{L}(\mathcal{M}|G,\mathbf{X},\mathbf{Y})$ over the inputs $(G,\mathbf{X},\mathbf{Y})$ is minimized. 

Furthermore, the problem of active learning for node classification is formally defined as follows. In each step $t$, given the graph $G$ and the attribute matrix $\mathbf{X}$, an active learning strategy $\mathcal{A}$ selects a node subset $S^{t} \subseteq U^{t-1}$ for querying the labels $\mathbf{y}_i$ for each node $i \in S^t$. 
After getting the new set of labelled nodes $S^t$, we obtain a set of all labelled nodes $L^t = S^t \bigcup L^{t-1}$ 
and a set of unlabelled nodes $U^{t} = U^{t-1} \setminus S^{t} $ prepared for the next iteration. 
Then $G$ and $\mathbf{X}$ with labels $\mathbf{y}_i$ of $i \in L^t$ are used as training data to train a model $\mathcal{M}$ at the end of each step $t$. 
We define the labelling budget $b$ as the total maximum number of nodes which are allowed to be labelled. 
The eventual goal is to maximize the performance of the node classification task under the budget $b$. 
To achieve this, active learning needs to carefully select nodes for labelling (i.e., choose $S^t$ at each step $t$).
The objective is to minimize the loss using all labelled nodes $L^t$ at each step $t$: 
\begin{equation}
	\min\limits_{L^t} \mathcal{L}(\mathcal{M},\mathcal{A}|G,\mathbf{X},\mathbf{Y})
\end{equation}

\section{Methodology}
\label{sec:methodology}
In this section, we introduce our active learning framework LSCALE for node classification in a top-down fashion. First, we describe the overview and the key idea of LSCALE.
Then we provide the details of each module used in LSCALE. 

The overview of our latent space clustering-based active learning framework is shown in Figure \ref{fig:framework}. 
The most important aspect of our framework is to design a suitable \textit{active learning latent space}, specifically designed for clustering-based active learning. 
Motivated by the limitations of previous methods, we design a latent space with two important properties: 
1) \textit{low label requirements:} the latent space representations can be learned effectively even with very few labels, by utilizing the representation power from all nodes (including unlabelled nodes) rather than only labelled nodes; 
2) \textit{informative distances:} in the latent space, distances between intra-class nodes should be smaller than distances between inter-class nodes. 
With the first property, LSCALE can learn effective node representations throughout the active learning process, even when very few labelled nodes have been acquired. 
The second property makes distances in our latent space informative with respect to active learning selection, 
ensuring that clustering-based active selection processes choose a diverse set of query points.

To satisfy the first property, we use an unsupervised learning method to learn unsupervised node representations $\mathbf{H} \in \mathbb{R}^{n\times d'}$ based on graphs and node attributes, where $d'$ is the dimension of representations. 
After obtaining $\mathbf{H}$, we design a linear distance-based classifier to generate output predictions.
In the classifier, we apply a learnable linear transformation on $\mathbf{H}$ to obtain hidden representations $\mathbf{Z}$.
The distances between nodes are calculated by a dynamic combination of both $\mathbf{Z}$ and $\mathbf{H}$, which satisfies the second desired property. 
Clustering is performed on the latent space using the distances to select informative nodes. 
We additionally propose an incremental clustering method to ensure that the newly selected nodes are not redundant with the previously selected nodes. 
In summary, the framework contains a few main components: 
\vspace{-\topsep}
\begin{itemize}
	\item an unsupervised graph learning method to generate unsupervised node representations.
	\item an active learning latent space with two aforementioned properties: 1) \emph{low label requirements}, 2) \emph{informative distances}. 
	\item an incremental clustering method to select data points as centroids, which we use as the nodes to be labelled, and prevent redundancy during node selection.
\end{itemize}
\subsection{Active Learning Latent Space} 
To facilitate clustering-based active learning, we need a latent space with two desired properties. Therefore, we propose a distance-based classifier for generating representations from supervised signals and a distance function to dynamically consider supervised and unsupervised representations simultaneously.  

\header \textbf{Distance-based classifier.}
In our framework, we design a novel distance-based classifier to ensure that distances between nodes in our latent space can facilitate active learning further.
Intuitively, a desired property of the latent space is that nodes from different classes should be more separated and nodes with the same class should be more concentrated in the latent space. Thus, it can help clustering-based active learning methods select representative nodes from different classes.
To achieve this, we first map unsupervised features $\mathbf{H}$ to another set of features $\mathbf{Z}$ by a linear transformation: 
\begin{equation}
	\mathbf{Z} = \mathbf{H}\mathbf{W}_c,
\end{equation}
where $\mathbf{W}_c \in \mathbb{R}^{d' \times l'}$ is the trainable linear transformation matrix. $l'$ is the dimension of latent representations. 
Then we define a set of learnable class representations ${\mathbf{c}_1, \mathbf{c}_2, ..., \mathbf{c}_K}$, where $K$ is the number of classes. The distance vector of node $i$ is defined as:
\begin{equation}
	\mathbf{a}_i = ||\mathbf{z}_i-\mathbf{c}_1||_2 \oplus ||\mathbf{z}_i-\mathbf{c}_2||_2 ... \oplus ||\mathbf{z}_i-\mathbf{c}_K||_2 \in \mathbb{R}^{K},
\end{equation}
where $\oplus$ is the concatenation operation and $||\cdot||_2$ is the $L_2$ norm.
The $j$-th element $\hat{y}_{ij}$ in the output prediction $\mathbf{\hat{y}}_i$ of node $i$ is obtained by the softmax function: 
\begin{equation}
	\hat{y}_{ij} = \textsf{softmax}(\mathbf{a}_i)_j = \frac{\exp(||\mathbf{z}_i-c_j||_2)}{\sum_{k=0}^K \exp(||\mathbf{z}_i-c_k||_2)}
\end{equation}

For training the classifier, suppose the labelled node set at step $t$ is $L^t$. The cross-entropy loss function for node classification over the labeled node set is defined as: 
\begin{equation}
	\mathcal{L} = - \frac{1}{|L^t|}\sum_{i \in L^t} \sum_{c=1}^{K} y_{ic} \ln \hat{y}_{ic},
\end{equation}
where $y_{ic}$ denotes the $c$-th element in the label vector $\mathbf{y}_i$.

With the guidance of labelled nodes and their labels via backpropagation, we can update the transformation matrix $\mathbf{W}_c$ 
and new features $\mathbf{Z}$ can capture the supervised information from labelled data. In addition, new features $\mathbf{Z}$ allow intra-class nodes more close and inter-class nodes more separate in the feature space. 
Through this distance-based classifier, the generated feature space allows the clustering-based active selection effectively select a diverse set of query nodes.

\header \textbf{Distance function.}
In LSCALE, the distance function determines the distances between nodes in the latent space for further clustering. 
We define our distance function as: 
\begin{equation}
	d(v_i, v_j) = ||g(\mathbf{X})_i - g(\mathbf{X})_j||_2, 
	\label{dis_func}
\end{equation}
where $g(\mathbf{X})$ 
is a mapping from node attributes $\mathbf{X}$ to new distance features.
As previous graph active learning methods do not effectively utilize the unlabelled nodes, we aim to take advantage of unsupervised learning features and supervised information from labelled data. 
To this end, we combine unsupervised learning features $\mathbf{H}$ and supervised hidden representations $\mathbf{Z}$ in the distance function. A straightforward way to combine them is using concatenation of $\mathbf{H}$ and $\mathbf{Z}$: $g(\mathbf{X}) = \mathbf{H} \oplus \mathbf{Z}$. Noted that $\mathbf{H}$ and $\mathbf{Z}$ are in different spaces and may have different magnitudes of row vectors. We define the distance features as follows: 
\begin{equation}
	g(\mathbf{X}) = \alpha \cdot \mathbf{H}' \oplus (1-\alpha) \cdot \mathbf{Z}',
\end{equation}
where $\mathbf{H}'$ and $\mathbf{Z}'$ are 
\textit{l}2-normalized $\mathbf{H}$ and $\mathbf{Z}$ respectively to make sure they have same Euclidean norms of rows. 
$\alpha$ can be treated as a parameter for controlling the dynamic combination of unsupervised features and supervised features. 

Intuitively, $\mathbf{Z}$ can be unstable in the early stages as there are relatively few labelled nodes in the training set. 
So, in the early stages, we would like to focus more on unsupervised features $\mathbf{H}$, which are much more stable than $\mathbf{Z}$. 
As the number of labelled nodes increases, the focus should be shifted to hidden representations $\mathbf{Z}$ in order to emphasize supervised information. 
Inspired by curriculum learning approaches \cite{curriculum-learning}, we set an exponentially decaying weight as follows: 
\begin{equation}
	\alpha = \lambda^{|L^{t}|},
	\label{eq: lambda}
\end{equation}
where $|L^{t}|$ is the number of labelled nodes at step $t$. $\lambda$ can be set as a number close to $1.0$, e.g., $0.99$. 
By using this dynamic combination of unsupervised learning features $\mathbf{H}$ and supervised hidden representations $\mathbf{Z}$, we eventually construct the latent space $g(\mathbf{X})$ which has the two important properties: 1) \emph{low label requirements:} it utilizes the representation power from all nodes including unlabelled nodes; 2) \emph{informative distances:} distances between nodes are informative for node selection.
Thus, the latent space can facilitate selecting diverse and representative nodes in the clustering module. 

Note that FeatProp \cite{wu2019active_featprop} uses propagated node attributes as representations for calculating distances. The propagated node attributes are fixed and not learnable throughout the whole active learning process, which makes the node selection less effective. 
In contrast, our latent space is learned based on signals from both labelled and unlabelled data.
In addition, it gradually shifts its focus to emphasize supervised signals as we acquire more labelled data.


\subsection{Clustering Module}
At each step, we use the K-Medoids clustering on our latent space to obtain cluster representatives. In K-Medoids, medoids are chosen from among the data points themselves, ensuring that they are valid points to select during active learning. So, after clustering, we directly select these medoids for labelling. This ensures that the chosen centers are well spread out and provide good coverage of the remaining data, which matches the intuition of active learning, since we want the chosen centers to help us classify as much as possible of the rest of the data.
At each step $t$, the objective of K-Medoids is:
\begin{equation}
	\sum_{i=1}^{n} \min_{j\in S^t} d(v_i, v_j) = \sum_{i=1}^{n} \min_{j\in S^t} ||g(\mathbf{X})_i - g(\mathbf{X})_j||_2
\end{equation}
Besides K-Medoids, common clustering methods used in the previous work are K-Means \cite{active_graph_embedding,ANRMAB} and K-Centers \cite{core-set}.
K-Means cannot be directly used for selecting nodes in active learning as it does not return real sample nodes as cluster representatives. 

\header \textbf{Incremental clustering.}
Despite these advantages of K-Medoids for active learning on graphs, a crucial drawback is that it is possible to select similar nodes for querying during multiple iterations. 
That is, newly selected nodes may be close to previously selected ones, making them redundant and hence worsening the performance of active learning.
The reason is that the clustering algorithm only generates the representative nodes in the whole representation space without the awareness of previously selected nodes. 
To overcome this problem, we design an effective incremental clustering algorithm for K-Medoids to avoid selecting redundant nodes.


In our incremental clustering method, the key idea is that fixing previous selected nodes as some medoids can force the K-Medoids algorithm to select additional medoids that are dissimilar with the previous ones. 
We illustrate our incremental clustering method in Algorithm \ref{algo:clustering}. 
\begin{algorithm}[t]
	\SetAlgoLined
	\LinesNumbered
	\KwIn{the set of previous labelled nodes $L^{t-1}$, \newline
		the set of unlabelled nodes $U^{t-1}$ as the pool, \newline
		the budget $b^t$ of the current step.}
	$k \leftarrow |L^{t-1}| + b^t$\;
	Randomly select $b^t$ nodes from $U^{t-1}$\;
	Set selected $b^t$ nodes and nodes in $L^{t-1}$ as k initial medoids\;
	Compute $d(v_i,v_j)$ for every node pair $(v_i,v_j)$ by Eq. (\ref{dis_func});
	
	\Repeat{all the medoids are not changed}{
		
		\ForEach{node $u \in U^{t-1}$}{Assign $u$ to the cluster with the closest medoid;
		}
		\ForEach{cluster $C$ with medoid $m$}{
			\If{$m \not\in L^{t-1}$}{
				Find the node $m'$ which minimize the sum of distances to all other nodes within $C$\;
				
				Update node $m'$ as the medoid of $C$\;
			}
		}
	}
	Construct selected node set $S^t$ using the medoids $m \not\in L^{t-1}$\;
	$L^t \leftarrow S^t \bigcup L^{t-1}$; $U^t \leftarrow U^{t-1} \setminus S^t$\;
	\KwRet{$L^t, U^t, S^t$}
	\caption{Incremental K-Medoids clustering}
	\label{algo:clustering} 
\end{algorithm}

After calculating the distances for every node pair (Line 4), incremental K-Medoids is conducted (Line 5 to Line 15). 
Compared to the original K-Medoids, the most important modification is that only clusters with a medoid, which is not in the previous labelled nodes set (i.e., $m \not\in L^{t-1}$), can update the medoid (Line 10-13). 
When all the medoids are the same as those in the previous iteration, the K-Medoids algorithm stops and keeps the medoids. For the medoids which are not the previous selected nodes, we put them in selected node set $S^t$, meanwhile we set labelled node set $L^t$ and unlabelled node set $U^t$ using $S^t$ accordingly.

\section{Experiments} 
\label{sec:Experiments}
The main goal of our experiments is to verify the effectiveness of our proposed framework LSCALE\footnote{The code can be found https://github.com/liu-jc/LSCALE}. 
We design experiments to answer the following research questions: 
\begin{itemize}
	\item \textbf{RQ1. Overall performance and effectiveness of unsupervised features}: How does LSCALE perform as compared with state-of-the-art graph active learning methods? Is utilizing unsupervised features also helpful for other clustering-based graph active learning methods? 
	\item \textbf{RQ2. Efficiency:} How efficient is LSCALE as compared with other methods? 
	\item \textbf{RQ3. Ablation study:} Are the designed dynamic feature combination and incremental clustering useful to improve the performance? How does our distance-based classifier affect the performance?
\end{itemize}

\textbf{Datasets.} To evaluate the effectiveness of LSCALE, we conduct the experiments on Cora, Citeseer \cite{cora_citeseer_data}, Pubmed \cite{pubmed_data}, Coauthor-CS (short as Co-CS) and Coauthor-Physics (short as Co-Phy) \cite{shchur2018pitfalls}. 
The first three are citation networks  while Co-CS and Co-Phy are two co-authorship networks.
We describe the datasets in detail and summarize the dataset statistics in Supplement B.1. 

\begin{table*}[t]
	 \small
	\caption{The averaged accuracies (\%) and standard deviations at different budgets on citation networks.}
	\centering
	\resizebox{0.99\linewidth}{!}{
	\begin{tabular}{@{}c|ccc|ccc|ccc@{}}
		\toprule
		\textbf{Dataset} & \multicolumn{3}{c|}{\textbf{Cora}} & \multicolumn{3}{c|}{\textbf{Citeseer}} & \multicolumn{3}{c}{\textbf{Pubmed}} \\ \midrule
		\textbf{Budget}      & 10 & 30 & 60 & 10 & 30 & 60 & 10 & 30 & 60 \\ \midrule
		Random      & 47.65$\pm$7.2   & 65.19$\pm$4.6   & 73.33$\pm$3.1   & 37.76$\pm$9.7   & 57.73$\pm$7.1   & 66.38$\pm$4.4   & 63.60$\pm$6.8   & 74.17$\pm$3.9   & 77.93$\pm$2.4     \\
		Uncertainty & 45.78$\pm$4.6   & 56.34$\pm$8.4   & 70.22$\pm$6.0   & 27.65$\pm$8.8   & 45.04$\pm$8.2   & 59.41$\pm$9.2   & 60.72$\pm$5.7   & 69.64$\pm$4.2   & 74.95$\pm$4.2     \\
		AGE         & 41.22$\pm$9.3   & 65.09$\pm$2.7   & 73.63$\pm$1.6   & 31.76$\pm$3.3   & 60.22$\pm$9.3   & 64.77$\pm$9.1   & 66.96$\pm$6.7   & 75.82$\pm$4.0   & 80.27$\pm$1.0    \\
		ANRMAB       & 30.43$\pm$8.2   & 61.11$\pm$8.8   & 71.92$\pm$2.3   & 25.66$\pm$6.6   & 47.56$\pm$9.4   & 58.28$\pm$9.2   & 57.85$\pm$8.7   & 65.33$\pm$9.6   & 75.01$\pm$8.4   \\
		FeatProp     & 51.78$\pm$6.7   & 66.49$\pm$4.7   & 74.70$\pm$2.7   & 39.63$\pm$9.2   & 57.92$\pm$7.2   & 66.95$\pm$4.2   & 67.33$\pm$5.5   & 75.08$\pm$3.2   & 77.60$\pm$1.9   \\
		GEEM  & 45.73$\pm$9.8 & 67.21$\pm$8.7 & 76.51$\pm$1.6 & 41.10$\pm$7.2 & 62.96$\pm$7.8 & \underline{70.82$\pm$1.2} & 64.38$\pm$6.7 &76.12$\pm$1.9 &79.10$\pm$2.3 \\ 
		\midrule
		DGI-Rand  & 62.55$\pm$5.8   & 73.04$\pm$3.8   & 78.36$\pm$2.6   & 54.46$\pm$7.6   & 67.26$\pm$4.0   & 70.24$\pm$2.4   & 73.17$\pm$3.8   & 78.10$\pm$2.8   & 80.28$\pm$1.6 \\ 
		FeatProp-D & 68.94$\pm$5.7& 75.47$\pm$2.9 & 77.64$\pm$2.0& 61.84$\pm$5.9& 66.99$\pm$3.6& 68.97$\pm$2.0& 73.50$\pm$4.7& 77.36$\pm$3.4& 78.54$\pm$2.3\\ 
		\midrule 
		LSCALE-D        & \underline{70.83$\pm$4.8}   & \underline{77.41$\pm$3.5}   & \underline{80.77$\pm$1.7}   & \textbf{65.60$\pm$4.7}   & \textbf{69.06$\pm$2.6}   & \textbf{70.91$\pm$2.2}   & \textbf{74.28$\pm$4.4}   & \underline{78.54$\pm$2.8}   & \underline{80.62$\pm$1.7}   \\
		LSCALE-M &\textbf{72.71$\pm$3.9} &\textbf{78.67$\pm$2.7} &\textbf{82.03$\pm$1.8} &\underline{64.24$\pm$4.8} &\underline{68.68$\pm$3.2} &70.34$\pm$1.9 &\underline{73.51$\pm$4.9} &\textbf{79.09$\pm$2.3} &\textbf{81.32$\pm$1.7} \\ 
		\bottomrule
	\end{tabular}
	}
	\label{tab:results}
\end{table*}

\begin{table}[t]
	\caption{The averaged accuracies (\%) and standard deviations at different budgets on co-authorship networks.}
	\centering
	\begin{tabular}{@{}c|ccc|ccc@{}}
		\toprule
		\textbf{Dataset} &  \multicolumn{3}{c|}{\textbf{Co-Phy}} & \multicolumn{3}{c}{\textbf{Co-CS}} \\ \midrule
		\textbf{Budget}      & 10 & 30 & 60 & 10 & 30 & 60 \\ \midrule
		Random      & 74.80   & 86.48   & 90.70$\pm$2.6   & 49.72   & 69.98   & 78.15$\pm$3.6   \\
		Uncertainty & 71.42   & 86.64   & 91.29$\pm$2.0   & 42.38   & 57.43   & 65.66$\pm$9.5   \\
		AGE        & 63.96   & 84.47   & 91.30$\pm$2.0   & 27.20   & 70.22   & 76.52$\pm$3.6   \\
		ANRMAB       & 68.47   & 84.19   & 89.35$\pm$4.2   & 43.48   & 69.98   & 75.51$\pm$2.4   \\
		FeatProp    & 80.23   & 86.83   & 90.82$\pm$2.6   & 52.45   & 70.83   & 76.60$\pm$3.9   \\
		GEEM & 79.24 & 88.58 & 91.56$\pm$0.5& 61.63 & 75.03 & 82.57$\pm$1.9\\
		\midrule
		DGI-Rand  & 82.81   & 90.35   & 92.44$\pm$1.5   & 64.07   & 78.63   & 84.28$\pm$2.7 \\ 
		FeatProp-D & 87.90 & 91.23 & 91.51$\pm$1.6& 67.37& 77.65& 80.33$\pm$2.5\\
		\midrule 
		LSCALE-D        & \textbf{90.38}   & \underline{92.75}   & \textbf{93.70$\pm$0.6}   & \textbf{73.07}   & \textbf{82.96}   & \textbf{86.70$\pm$1.7}   \\
		LSCALE-M & \underline{90.28} & \textbf{92.89} & \underline{93.05$\pm$0.7} &\underline{71.16} & \underline{81.82} & \underline{85.79$\pm$1.6} \\ 
		\bottomrule
	\end{tabular}
	\label{tab:small_results}
\end{table}

\textbf{Baselines.}
In the experiments, to show the compatibility with different unsupervised learning methods, we use two variants LSCALE-DGI and LSCALE-MVGRL, which use DGI \cite{DGI} and MVGRL \cite{ICML-MVGRL} as the unsupervised learning method, respectively. 
To demonstrate the effectiveness of LSCALE, we compare two variants with the following representative active learning methods on graphs. \textbf{Random}: select the nodes uniformly from the unlabelled node pool; \textbf{Uncertainty}: select the nodes with the max information entropy according to the current model. \textbf{AGE} \cite{active_graph_embedding} constructs three different criteria based on graph neural networks to choose a query node. Combining these different criteria with time-sensitive variables to decide which nodes to selected for labelling. \textbf{ANRMAB} \cite{ANRMAB} proposes a multi-arm-bandit mechanism to assign different weights to the different criteria when constructing the score to select a query node. \textbf{FeatProp} \cite{wu2019active_featprop} performs the K-Medoids clustering on the propagated features obtained by simplified GCN \cite{SGC} and selects the medoids to query their labels. \textbf{GEEM} \cite{ICML-regol20-active}: inspired by error reduction, it uses simplified GCN \cite{SGC} to select the nodes by minimizing the expected error.

As suggested in \cite{active_graph_embedding,ANRMAB,wu2019active_featprop}, AGE, ANRMAB, FeatProp, Random, and Uncertainty use GCNs as the prediction model, which is trained after receiving labelled nodes at each step. 
GEEM uses the simplified graph convolution (SGC) \cite{SGC} as the prediction model as mentioned in \cite{ICML-regol20-active}.

\subsection{Experimental Setting}
We evaluate LSCALE-DGI, LSCALE-MVGRL, and other baselines on node classification task with a transductive learning setup, following the experimental setup as in \cite{ANRMAB,wu2019active_featprop} for a fair comparison.

\header \textbf{Dataset splits.}
For each citation dataset, we use the same testing set as in \cite{semi_GCN}, which contains 1000 nodes. 
For coauthor datasets, we randomly sample 20\% nodes as the testing sets.
From the non-testing set in each dataset, we randomly sample 500 nodes as a validation set and fix it for all the methods to ensure a fair comparison. 


\header \textbf{Experiment procedure.} In the experiments, we set the budget sizes differently for different datasets and we focus on the ``batched'' multi-step setting as in \cite{core-set,wu2019active_featprop}.  
Each active learning method is provided a small set of labelled nodes as an initial pool. 
As in \cite{wu2019active_featprop}, we randomly select 5 nodes regardless of the class as an initial pool.
The whole active learning process is as follows: (1) we first train the prediction model with initial labelled nodes. (2) we use the active learning strategy to select new nodes for labelling and add them to the labelled node pool; (3) we train the model based on the labelled nodes again. We repeat Step (2) and Step (3) until the budget is reached and train the model based on the final labelled node pool. 
For clustering-based methods (i.e., FeatProp and LSCALE), 10 nodes are selected for labelling in each iteration as these methods depend on selecting medoids to label. 

\header \textbf{Hyperparameter settings.} For hyperparameters of other baselines, we set them as suggested in their papers.
We specify hyperparameters of our methods in Supplement B.2.
\subsection{Performance Comparison (RQ1)}
\header \textbf{Overall comparison.} We evaluate the performance by using the averaged classification accuracy. We report the results over 20 runs with 10 different random data splits.
In Table \ref{tab:results} and  \ref{tab:small_results}, we show  accuracy scores of different methods when the number of labelled nodes is less than 60. 
Analysing Table \ref{tab:results} and \ref{tab:small_results}, we have the following observations:
\begin{itemize}
	\item  In general, our methods LSCALE-DGI (short as LSCALE-D) and LSCALE-MVGRL (short as LSCALE-M) significantly outperform the baselines on the varying datasets, while they provide relatively lower standard deviations on most datasets.
	In particular, when the total budget is only 10, LSCALE-M provides remarkable improvements compared with GEEM by absolute values 26.9\%, 23.3\%, 11.5\%, on Cora, Citeseer, and Co-CS respectively. 
	\item With the budget size less than 30, the Uncertainty baseline always performs worse than the Random baseline for all datasets. Meanwhile,  AGE and ANRMAB do not have much higher accuracies on most datasets compared with the Random baseline. 
	Both of the above results indicate that GCN representations, which are used in AGE, ANRMAB and the Uncertainty baseline for selecting nodes, are inadequate when having only a few labelled nodes. 
	\item GEEM generally outperforms other baselines on all datasets, which might be attributed to its expected error minimization scheme. However, the important drawback of expected error minimization is the inefficiency, which we show later in Sec \ref{sec: efficiency comparison}. 
\end{itemize}
Supplement B.3 shows more results about how accuracy scores of different methods change as the number of labelled nodes increases. 
Supplement B.4 demonstrates how different hyperparameters affect the performance.

\header \textbf{Effectiveness of utilizing unsupervised features.}
Existing works overlook the information in unlabelled nodes, whereas LSCALE utilizes unsupervised features by using unsupervised learning on all nodes (including unlabelled ones). As we argue before, the information in unlabelled nodes is useful for active learning on graphs. 
To verify the usefulness, we design two additional baselines as follows: 
\begin{itemize}
	\item \textbf{FeatProp-DGI}: It replaces the propagated features with unsupervised DGI features in FeatProp to select nodes for labelling.
	\item \textbf{DGI-Rand}: It uses unsupervised DGI features and randomly selects nodes from the unlabelled node pool to label. For simplicity, it trains a simple logistic regression model with DGI features as the prediction model.
\end{itemize}
Regarding the effectiveness of unsupervised features, from Table \ref{tab:results} and \ref{tab:small_results}, we have the following observations: 
\begin{itemize}
	\item On all datasets, FeatProp-DGI (short as FeatProp-D) consistently outperforms FeatProp, which indicates unsupervised features are useful for other clustering-based graph active learning approaches besides our framework.
	\item DGI-Rand also achieves better performance compared with AGE, ANRMAB, and GEEM, especially when the labelling budget is small (e.g., 10). This verifies again that existing approaches do not fully utilize the representation power in unlabelled nodes.
	\item DGI-Rand outperforms FeatProp-D when the labelling budget increases to 60. This observation shows that FeatProp-D cannot effectively select informative nodes in the late stage, which can be caused by redundant nodes selected in the late stage.  
	\item While DGI-Rand and FeatProp-D use the representation power in unlabelled nodes, they are still consistently outperformed by LSCALE-D and LSCALE-M, which verifies the superiority of our framework. 
\end{itemize}
\subsection{Efficiency Comparison (RQ2)} \label{sec: efficiency comparison}
\begin{table}[t]
	\caption{The total running time of different models.}
    \centering
	\begin{tabular}{@{}c|ccccc@{}}
		\toprule
		\textbf{Method} & Cora    & Citeseer & Pubmed & Co-CS  & Co-Phy   \\ \midrule
		AGE             & 208.7s   & 244.1s    & 2672.8s & 6390.5s & 745.5s  \\
		ANRMAB          & 201.8s    & 231.5s     & 2723.3s   & 6423.5s  & 767.1s    \\
		FeatProp        & 16.5s    & 16.7s     & 58.7s   & 169.2s  & 336.4s    \\
		GEEM            & 3.1hr & 5.2hr  & 1.8hr & 52.5hr       & 46.2hr \\
		LSCALE-D      & 13.1s    & 15.6s     & 53.4s   & 59.8s   & 131.3s    \\ \bottomrule
	\end{tabular}
	\label{tab: efficiency}
\end{table}
We empirically compare the efficiency of LSCALE-D with that of four state-of-the-art methods (i.e., AGE, ANRMAB, FeatProp, and GEEM). Table \ref{tab: efficiency} shows the total running time of these models on different datasets. From Table \ref{tab: efficiency}, GEEM has worst efficiency as it trains the simplied GCN model $n\times K$ times ($K$ is the number of classes) for selecting a single node.
FeatProp and LSCALE-D are much faster than the other methods. The reason is that FeatProp and LSCALE-D both select several nodes in a step and train the classifier once for this step, whereas AGE, ANRMAB and GEEM all select a single node once in a step. 
Comparing LSCALE-D and FeatProp, LSCALE-D requires less time as the clustering in LSCALE-D is performed in the latent space where the dimension is less than that in the original attribute space used in FeatProp. 

\subsection{Ablation Study (RQ3)}

\begin{figure}
	\centering
	\begin{subfigure}[b]{0.48\columnwidth}
		\centering
		\includegraphics[width=\linewidth]{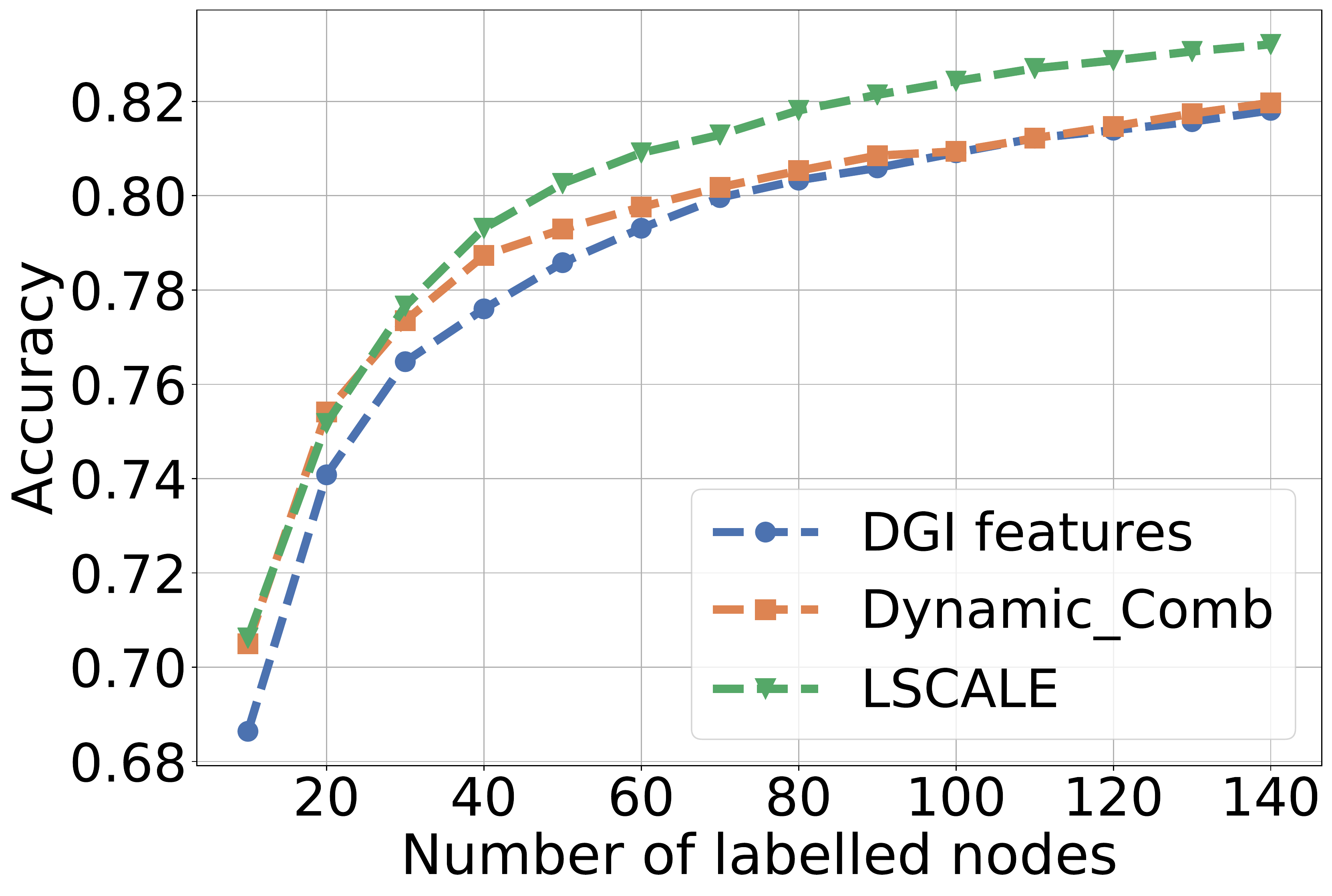}
		\caption{Cora}
		\label{fig:1}
	\end{subfigure}
	\begin{subfigure}[b]{0.48\columnwidth}
		\centering
		\includegraphics[width=\linewidth]{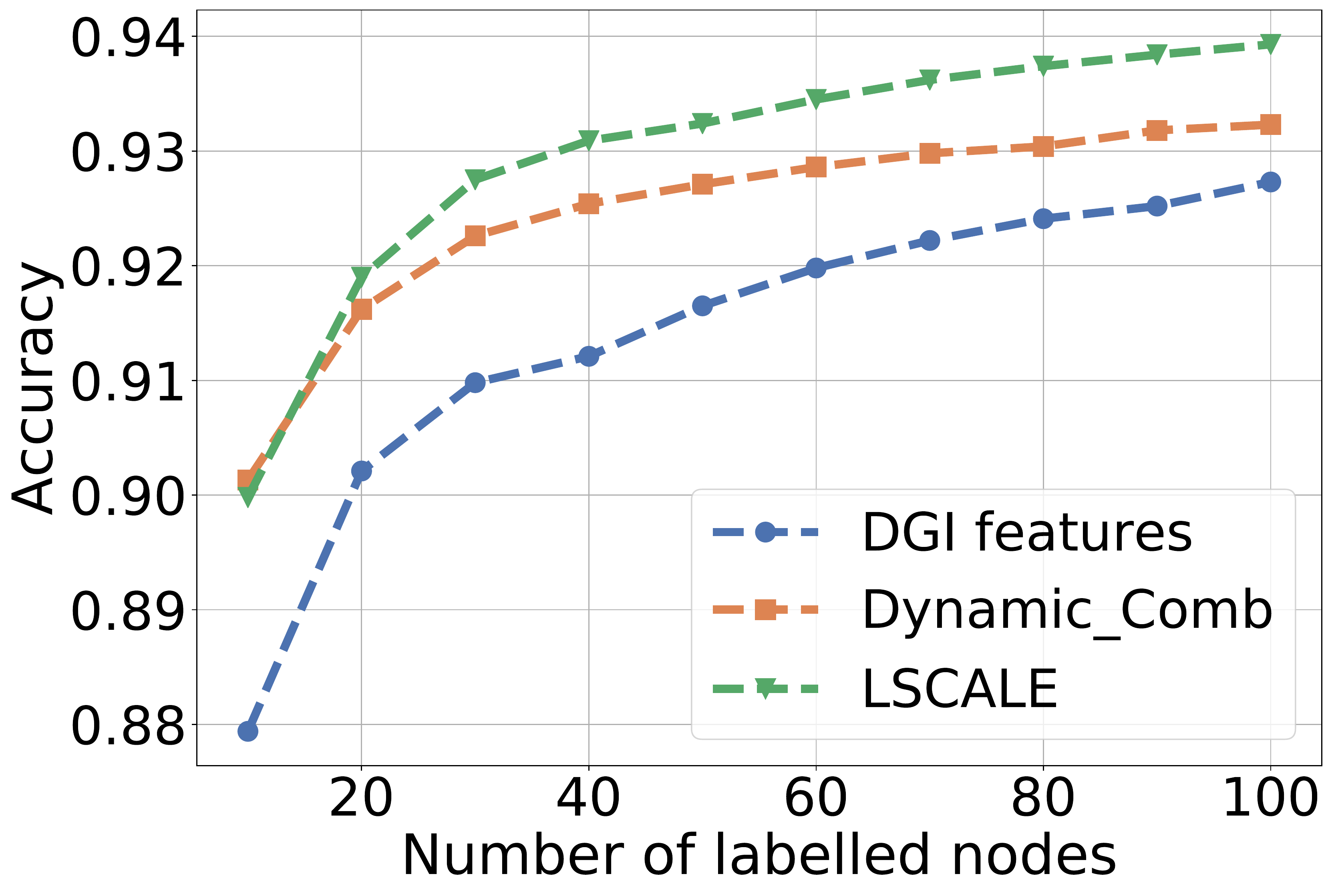}
		\caption{Co-Phy}
		\label{fig:2}
	\end{subfigure}
	\caption{Ablation study: dynamic feature combination and incremental clustering.}
	\label{fig:albation}
\end{figure}

\header \textbf{Effectiveness of dynamic feature combination and incremental clustering.}
\label{sec: ablation study}
We conduct an ablation study to evaluate the contributions of two different components in our framework: dynamic feature combination and incremental clustering. The results are shown in Figure \ref{fig:albation}.
{DGI features} is the variant without either dynamic combination or incremental clustering, and it only uses features obtained by DGI as distance features for the K-Medoids clustering algorithm. 
{Dynamic\_Comb} uses the dynamic combination to obtain distance features for clustering. 
{LSCALE} is the full version of our variant with dynamic feature combination and incremental clustering. 
It is worth noting that {DGI features} can be considered as a simple method utilizing unsupervised features. 
Analysing Figure \ref{fig:albation}, we have the following observations: 
\begin{itemize}
    \item {Dynamic\_Comb} generally provides better performance than {DGI features}, which shows the effectiveness of our dynamic feature combination for distance features. 
    \item {LSCALE} and {Dynamic\_Comb} provide no much different performance when the number of labelled nodes is relatively low. However, {LSCALE} gradually outperforms {Dynamic\_Comb} as the number of labelled nodes increases. This confirms that incremental clustering can select more informative nodes by avoiding redundancy between nodes selected at different steps. 
\end{itemize}
\begin{figure*}[t]
	\centering
	\begin{subfigure}[b]{0.325\textwidth}
		\centering
		\includegraphics[width=\textwidth]{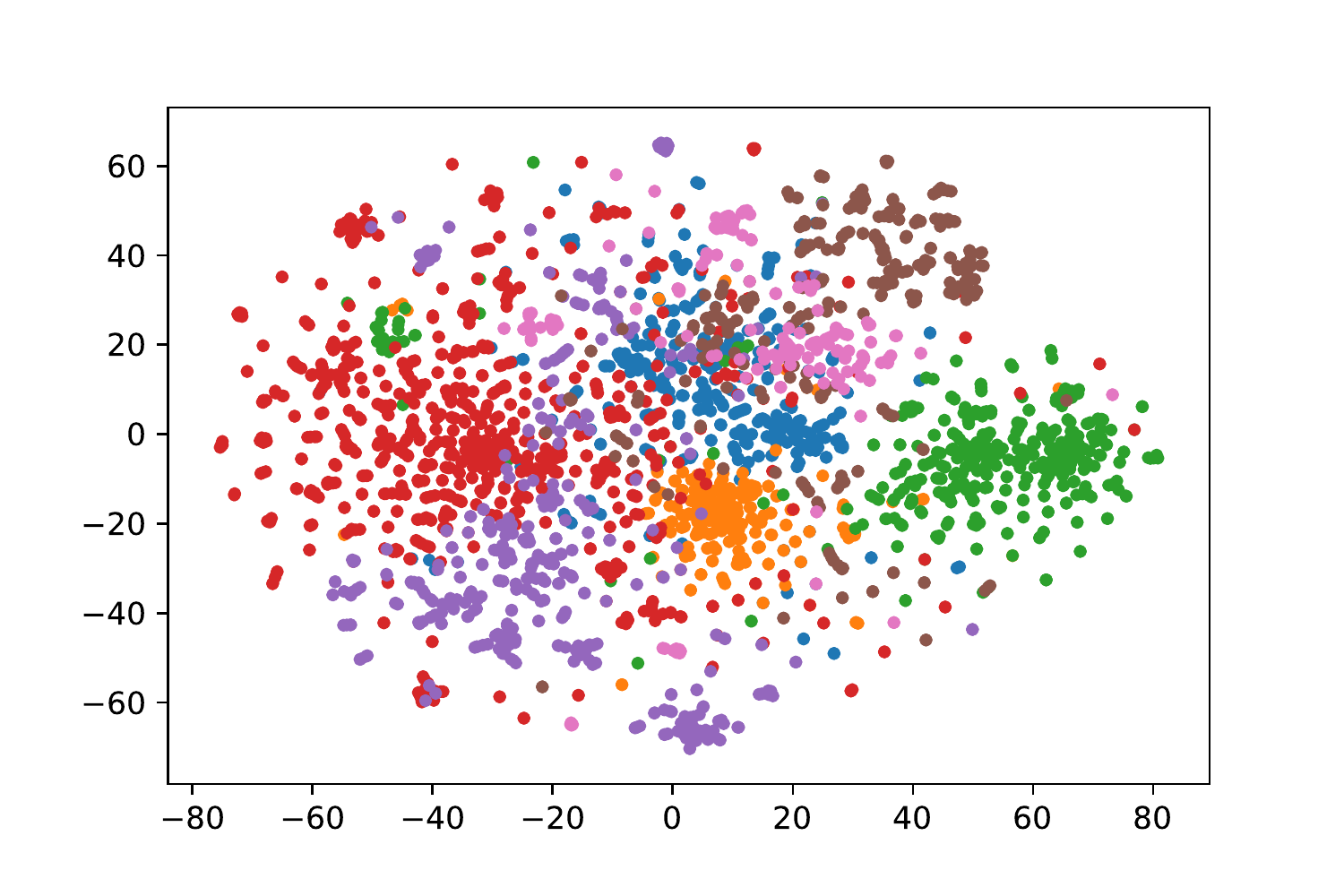}
		\caption{FeatProp}
	\end{subfigure}
	\begin{subfigure}[b]{0.325\textwidth}
		\centering
		\includegraphics[width=\textwidth]{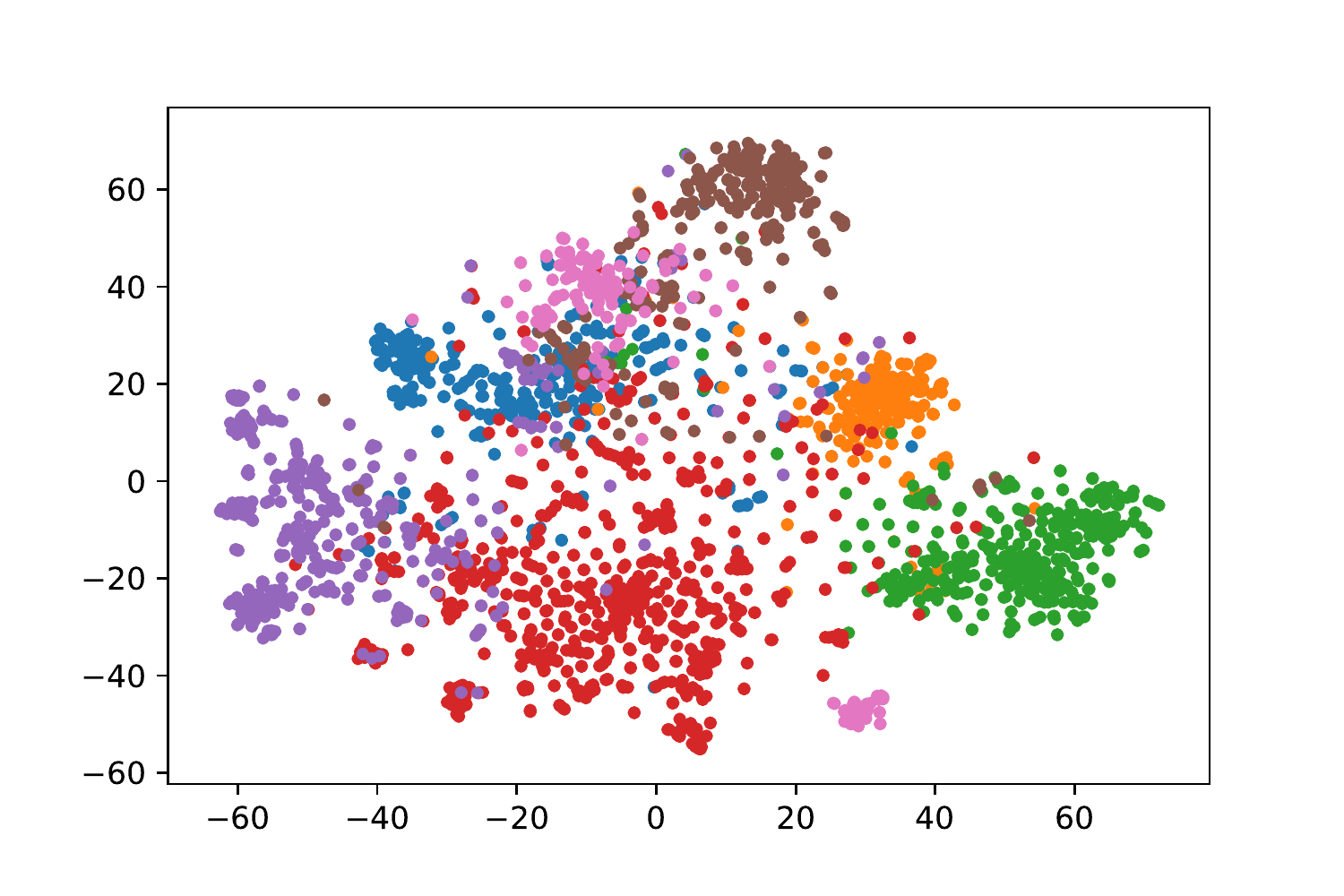}
		\caption{DGI}
	\end{subfigure}
	\begin{subfigure}[b]{0.325\textwidth}
		\centering
		\includegraphics[width=\textwidth]{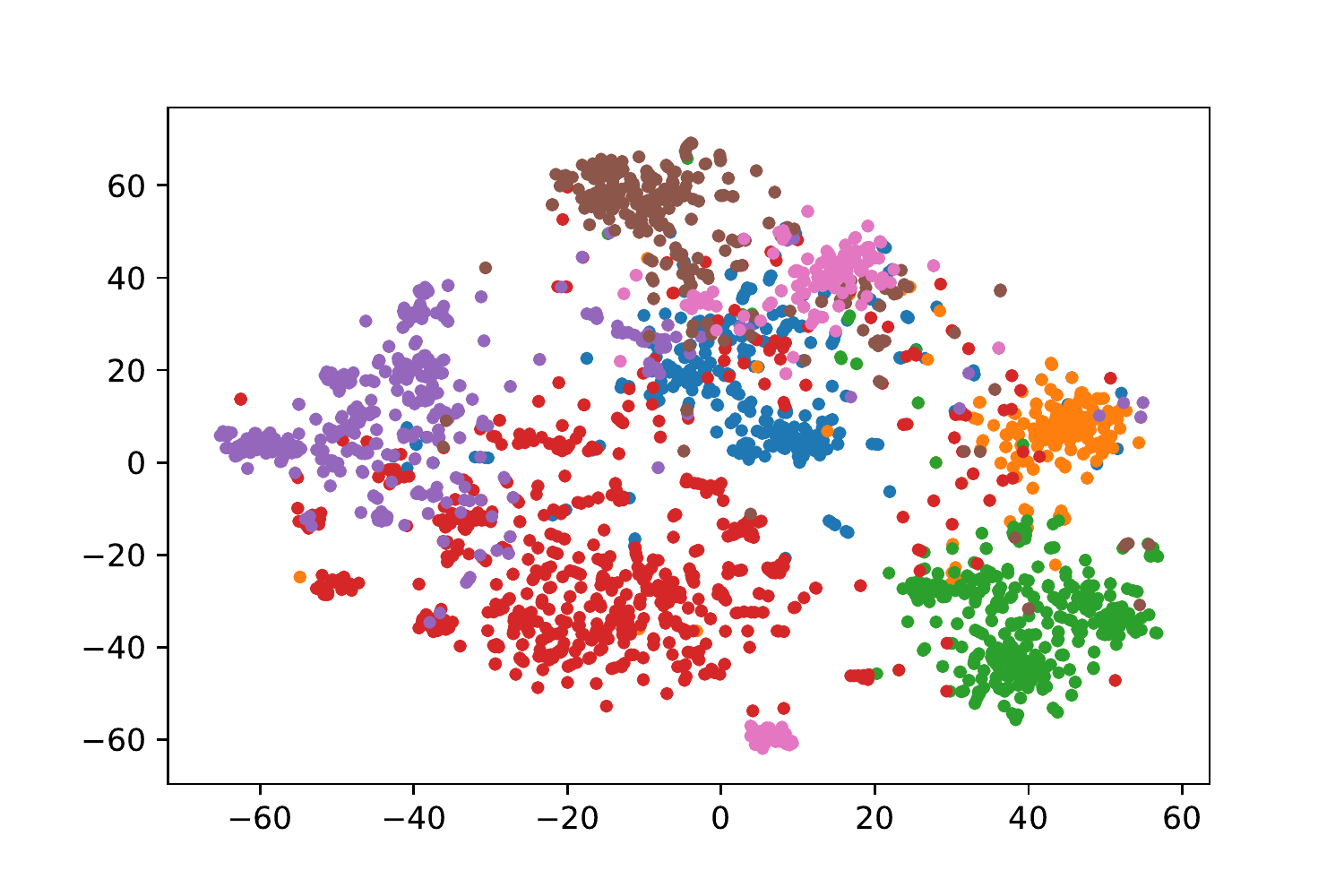}
		\caption{LSCALE-DGI}
	\end{subfigure}
	\caption{t-SNE visualization of different distance features on Cora dataset. Colors denote the ground-truth class labels. Our distance features have clearer separations between different classes.}
	\label{fig:tsne}
\end{figure*}
In summary, the results verify the effectiveness and necessity of dynamic combination and incremental clustering.

\header \textbf{Effectiveness of our distance features $g(\mathbf{X})$.}
Furthermore, we also qualitatively demonstrate the effectiveness of our distance feature $g(\mathbf{X})$. 
Figure \ref{fig:tsne} shows t-SNE visualizations \cite{tsne} of FeatProp features, DGI features, and the distance features of LSCALE-DGI. 
The distance features are obtained by dynamically combining DGI features and supervised hidden representations on 20 labelled nodes. 
Recall that FeatProp uses propagated node attributes as distance features and DGI features are learned using an unsupervised method with unlabelled data. 
Compared with others, the distance features used in LSCALE-DGI have clearer boundaries between different classes, which satisfies our second desired property (\textit{informative distances}) and further facilitates selecting informative nodes in the clustering algorithm.

\begin{table}[t]
	\caption{Accuracy (\%) comparison of LSCALE-DGI with different classifiers. The budget size is $20 \cdot c$  ($c$ is the number of classes).}
    \centering
	\begin{tabular}{@{}c|ccccc@{}}
		\toprule
		Classifier                      & Cora  & Citeseer & Pubmed & Co-CS & Co-Phy \\ \midrule
		GCN Classifier & 81.83 & 71.24    & 80.03  & 87.28 & 93.34  \\
		Distance Classifier & 83.23 & 72.30    & 80.62  & 89.25 & 93.97  \\ \bottomrule
	\end{tabular}
	\label{tab: diff classifier}
\end{table}
\begin{table}[t]
\centering
\caption{Accuracy (\%) comparison of FeatProp with different classifiers. }
\begin{tabular}{@{}c|ccccc@{}}
\toprule
Classifier                      & Cora  & Citeseer & Pubmed & Co-CS & Co-Phy \\ \midrule
GCN Classifier   & 80.50 & 72.04    & 77.65  & 83.49 & 93.06  \\
Distance Classifier & 80.66 & 72.14    & 77.88  & 84.32 & 93.28  \\ \bottomrule
\end{tabular}
\label{tab: diff classifier FeatProp}
\end{table}

\header \textbf{Effectiveness of  distance-based classifier.}
We design a distance-based classifier in LSCALE to ensure that distances are informative in the active learning latent space. 
To demonstrate the effectiveness of the distance-based classifier, we replace it with a GCN classifier and show the comparison in Table \ref{tab: diff classifier}. 
With the distance-based classifier, LSCALE can achieve better performance than that with a GCN classifier. 
This comparison shows the effectiveness of the designed distance-based classifier. 

To further investigate whether the distance-based classifier is also effective for other active learning methods, we change the GCN classifier to our proposed distance-based classifier for FeatProp and present the comparison in Table \ref{tab: diff classifier FeatProp}.
From Table \ref{tab: diff classifier FeatProp}, we note that FeatProp with our distance-based classifier has slightly better performance compared with FeatProp with GCN classifier on all the datasets. 
This observation indicates that our distance-based classifier is also effective for other clustering-based active learning methods.

\section{Related Work} 
\label{sec: related_works}
\textbf{Active Learning on Graphs}. For active learning on graphs, early works without using graph representations are proposed in \cite{bilgic2010active_networked_data,KDD_Active_2011,berberidis2018data_active_graph_cognizant}, where the graph structure is used to train the classifier and calculate the query scores for selecting nodes.
More recent works \cite{COLT_active_2015,CVPR_Active} study non-parametric classification models with graph regularization for active learning with graph data.

Recent works \cite{active_graph_embedding,ANRMAB,wu2019active_featprop,ICML-regol20-active} utilize graph convolutional neural networks (GCNs) \cite{semi_GCN}, which consider the graph structure and the learned embeddings simultaneously. 
AGE \cite{active_graph_embedding} design an active selecting strategy based on a weighted sum of three metrics considering the uncertainty, the the graph centrality and the information density. 
Improving upon the weight assignment mechanism, ANRMAB \cite{ANRMAB} designs a multi-armed bandit method with a reward scheme to adaptively assign weights for the different metrics. 

Besides the metric-based active selection on graphs, FeatProp \cite{wu2019active_featprop} uses a clustering-based active learning method, which calculates the distances between nodes based on representations of a simplified GCN model \cite{SGC} and conducts a clustering algorithm (i.e., K-Medoids) for selecting representative nodes. 
A recent method GEEM \cite{ICML-regol20-active} uses a simplified GCN \cite{SGC} for prediction and maximizes the expected error reduction to select informative nodes to label. 
Rather than actively selecting nodes and training/testing on a single graph, \cite{transferable-active-learning} learns a selection policy on several labelled graphs via reinforcement learning and actively selects nodes using that policy on unlabelled graphs. 
\cite{li2020seal,madhawa2020metal} use adversarial learning and meta learning approaches for active learning on graphs. However, even with relatively complicated learning methods, their performance are similar with AGE \cite{active_graph_embedding} and ANRMAB \cite{ANRMAB}. 
\cite{ActiveHNE} investigates active learning on heterogeneous graphs.
\cite{NEURIPS2021_eb86d510} considers noisy oracle setting where labels obtained by an oracle can be incorrect.  
In this work, we focus on the homogeneous single-graph setting like in \cite{active_graph_embedding,ANRMAB,wu2019active_featprop,ICML-regol20-active}. To tackle limitations of previous work on this setting, we have presented an effective and efficient framework that can utilize the representation power in unlabelled nodes and achieve better performance under the same labelling budget. 


\section{Conclusion}
\label{sec:conclusion}
In this paper, we focus on active learning for node classification on graphs and argue that existing methods are still less than satisfactory as they do not fully utilize the information in unlabelled nodes. 
Motivated by this, we propose LSCALE, a latent space clustering-based active learning framework, which uses a latent space with two desired properties for clustering-based active selection.
We also design an incremental clustering module to minimize redundancy between nodes selected at different steps. 
Extensive experiments demonstrate that our method provides superior performance over the state-of-the-art models. 
Our work points out a new possibility for active learning on graphs, which is to better utilize the information in unlabelled nodes by designing a feature space more suitable for active learning. 
Future work could propose new unsupervised methods which are more integrated with active learning process and enhance our framework further. 

\section*{Acknowledgements}
This paper is supported by the Ministry of Education, Singapore (Grant Number MOE2018-T2-2-091) and A*STAR, Singapore (Number A19E3b0099).

\bibliographystyle{splncs04}
\bibliography{ref}
\appendix
\onecolumn
{
\centering
\LARGE \bf Supplementary Materials \par
}
\section{Symbols}
\begin{table}[h]
\caption{Major symbols and definitions}
    \centering
    \begin{tabular}{@{}l|l@{}}
         \toprule
         \textbf{Symbol} &  \textbf{Definition} \\
         \midrule
         $G = (V, E)$ & Graph $G$ with node set $V$ and edge set $E$ \\
         $n,d$ & The number of nodes and the number of attributes in $G$\\
          $\mathbf{A}, \mathbf{D}$ & The adjacency matrix and the degree matrix  of graph $G$ \\ 
         \midrule
         $S^t$ & The set of selected nodes at step $t$ \\
         $U^t$ & The set of unlabelled nodes at step $t$ \\ 

         $L^t$ & The set of labelled nodes at step $t$ \\
         \midrule
         $\mathbf{X}, \mathbf{Y}$ & Node attributes and labels \\

         $\mathbf{\hat Y}$ & The model predictions \\
         $d(v_i, v_j)$ & The distance between node $v_i$ and $v_j$ for clustering \\
         \bottomrule
    \end{tabular}
    \label{tab:notations}
\end{table}
\section{More on Experiments}
\subsection{Datasets}
\label{supplement: datasets}
we conduct the experiments on Cora, Citeseer \cite{cora_citeseer_data}, Pubmed \cite{pubmed_data}, Coauthor-CS (short as Co-CS) and Coauthor-Physics (short as Co-Phy) \cite{shchur2018pitfalls}. 
The first three are citation networks, which are undirected networks and contain unweighted edges among different publications. Node attributes in Cora and Citeseer are provided as bag-of-words features, while node attributes of Pubmed are TF/IDF weighted word vectors. 
Co-CS and Co-Phy are two co-authorship networks where nodes are authors, which are connected by an edge if they coauthor a paper. Node attributes in these datasets are bag-of-words encoded paper keywords.
We summarize the dataset statistics for different datasets in Table \ref{Dataset statistic}. 
\begin{table}[t]
	\centering
	\caption{Dataset Statistics}
	\begin{tabular}{c|cccc}
		\toprule
		\textbf{Dataset}  &  \textbf{\#Nodes} &  \textbf{\#Edges} &  \textbf{\#Classes} &  \textbf{\#Attributes} \\ \midrule
		Cora     & 2,708    & 5,429    & 7          & 1,433                        \\ 
		Citeseer & 3,327    & 4,732    & 6          & 3,703                          \\ 
		Pubmed   & 19,717   & 44,338   & 3          & 500                            \\ 
		Co-CS & 18,333    & 81,894    & 15    & 6,805                           \\ 
		Co-Phy & 34,493   & 247,962   & 5 & 8,415                            \\ 
		\bottomrule
	\end{tabular}
	\label{Dataset statistic}
\end{table}
\subsection{Hyperparameter settings}
\label{supplement: hyperparameters}
For hyperparameters of other baselines, e.g., the number of layers of GCNs, the number of hidden units of GCNs, the optimizer, the learning rate, we set them as suggested in their papers.
For the embedding dimensionality $d'$ and other hyperparameters of corresponding unsupervised methods in LSCALE-DGI and LSCALE-MVGRL, we set them as suggested in the corresponding paper \cite{DGI,ICML-MVGRL}.
For the dimensionality $l'$ of hidden representation $Z$, we set as $100$ for all datasets. 
By default, We set $\lambda=0.99$ to determine $\alpha$  at different steps. 
To train the classifier in LSCALE, we use the Adam \cite{Adam} optimizer with the learning rate $0.2$ and the weight decay $5\times 10^{-6}$ for maximum 300 epochs and early stopping with a window size of 10.
\subsection{More Experimental Results}
\label{appendix: more results}
\begin{figure*}[t]
     \centering
     \begin{tabular}{c}
        \begin{subfigure}[b]{0.91\textwidth}
         \centering
         \includegraphics[width=\textwidth]{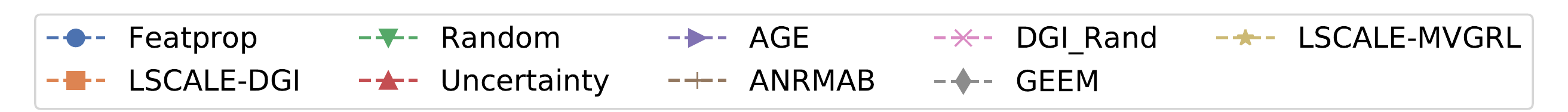}
         \end{subfigure}  \\
        \begin{subfigure}[b]{0.45\textwidth}
         \centering
         \includegraphics[width=\textwidth]{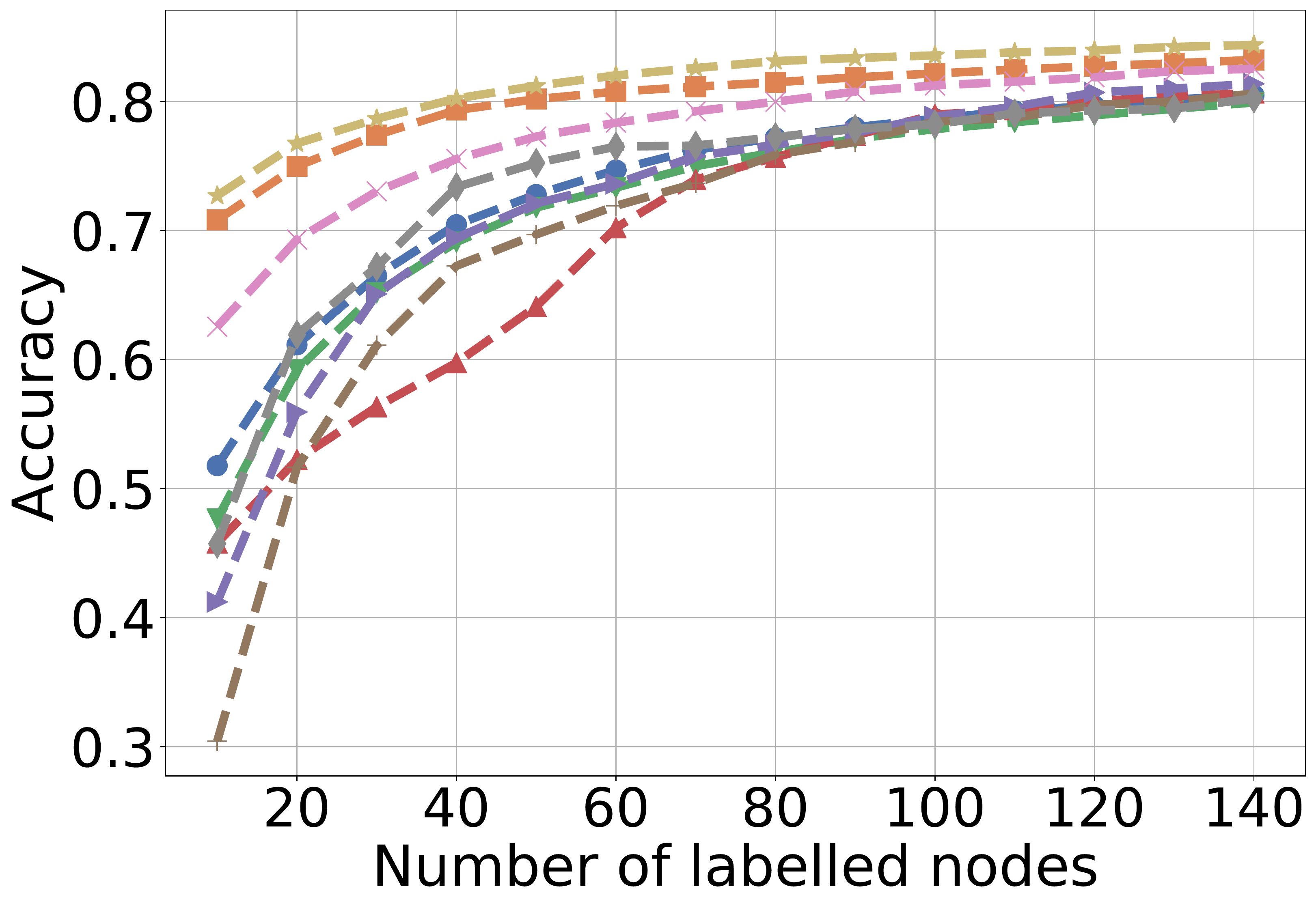}
         \caption{Cora}
         \end{subfigure}
         \begin{subfigure}[b]{0.45\textwidth}
             \centering
             \includegraphics[width=\textwidth]{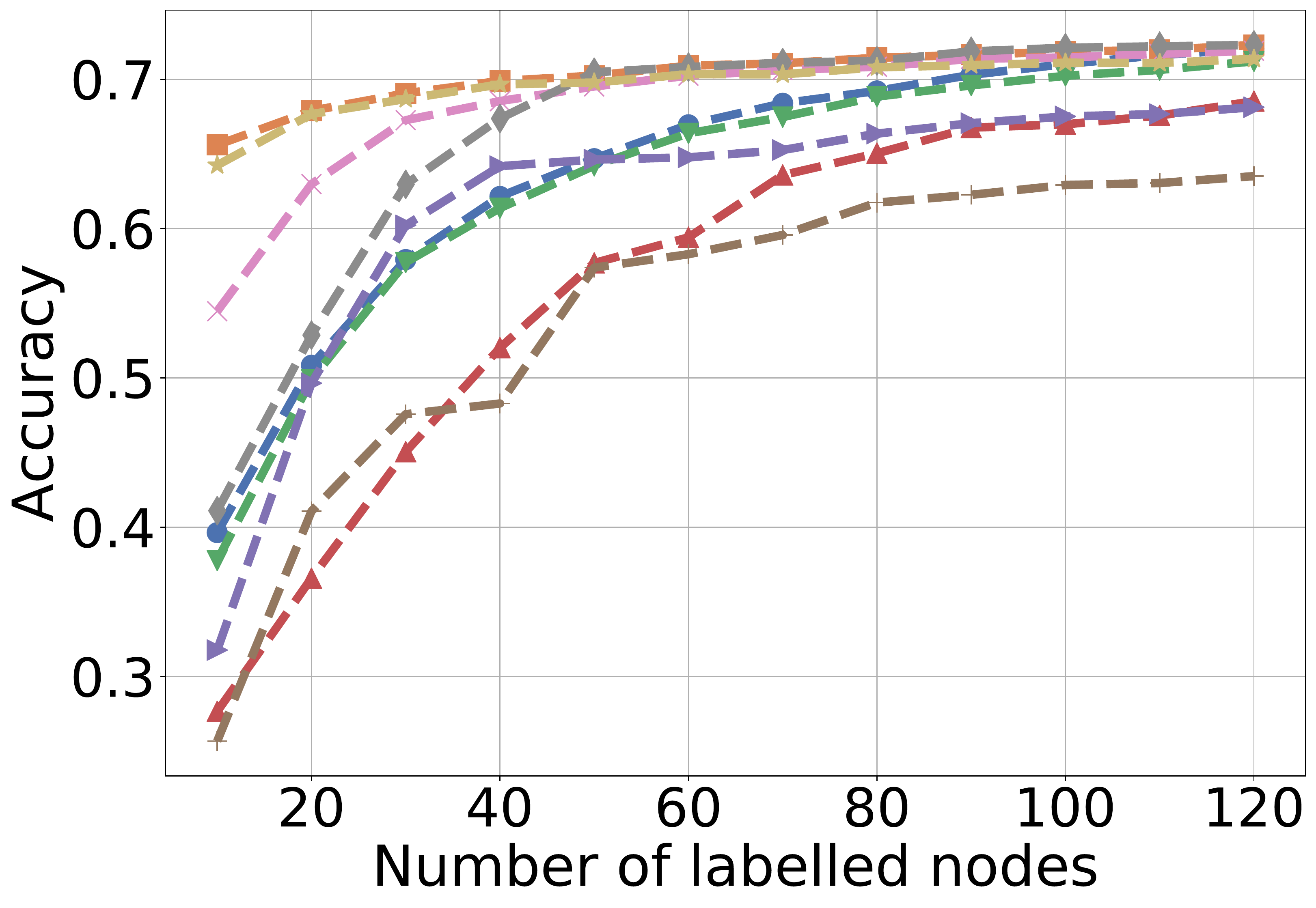}
             \caption{Citeseer}
         \end{subfigure} \\
         \begin{subfigure}[b]{0.45\textwidth}
             \centering
             \includegraphics[width=\textwidth]{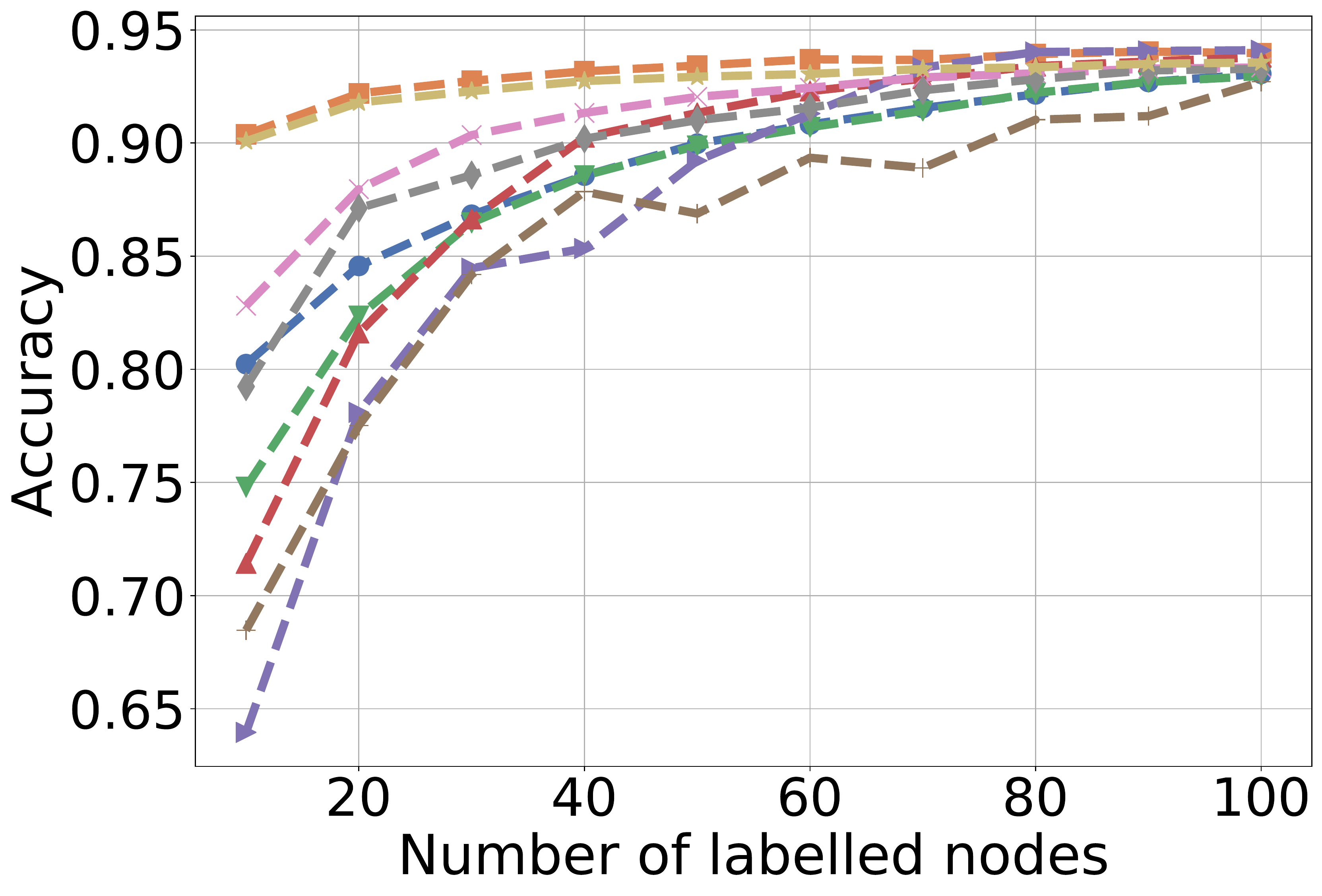}
             \caption{Co-Phy}
         \end{subfigure}
         \begin{subfigure}[b]{0.45\textwidth}
             \centering
             \includegraphics[width=\textwidth]{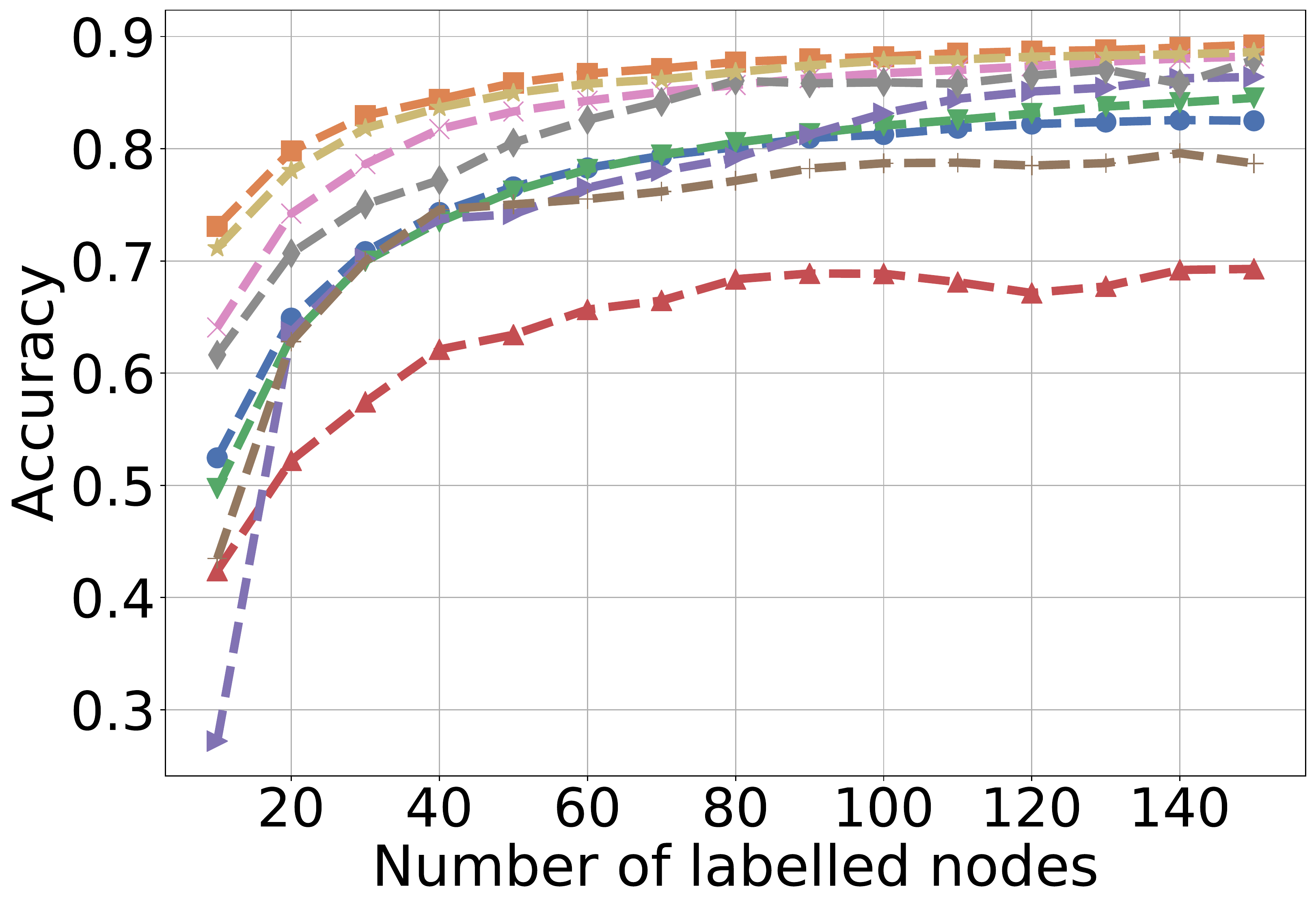}
             \caption{Co-CS}
         \end{subfigure}
     \end{tabular}
     \vspace{-3mm}
     \caption{Performance comparison of different active learning algorithms.}
     \label{fig:results}
     \vspace{-3mm}
\end{figure*}

In Figure \ref{fig:results}, we focus on how accuracy scores of different methods change as the number of labelled nodes increases on Cora, Citeseer, Co-Phy, and Co-CS. 
We omit Pubmed in Figure \ref{fig:results} because of space limitations and also the similar performance as other datasets.  
Comparing AGE and FeatProp on Cora, Co-Phy, and Co-CS, FeatProp initially outperforms AGE. 
However, AGE outperforms FeatProp as more labelled node are obtained, showing that supervised information is indeed helpful for effectively selecting informative nodes. 
Recall that FeatProp only uses propagated node attributes without considering supervised information during the node selection process, which makes it less effective as the number of labelled nodes increases on some datasets. 
GEEM consistently outperforms other state-of-the-art methods on Citeseer and Co-CS. 
Our two variants LSCALE-DGI and LSCALE-MVGRL generally outperform other methods under different budgets on all datasets, thanks to the proposed dynamic combination of unsupervised learning features and supervised hidden representations. 

\subsection{Parameter Study}
\label{appendix: Param_study}
\begin{figure}
    \centering
  \begin{subfigure}[b]{0.44\columnwidth}
    \centering
    \includegraphics[width=\linewidth]{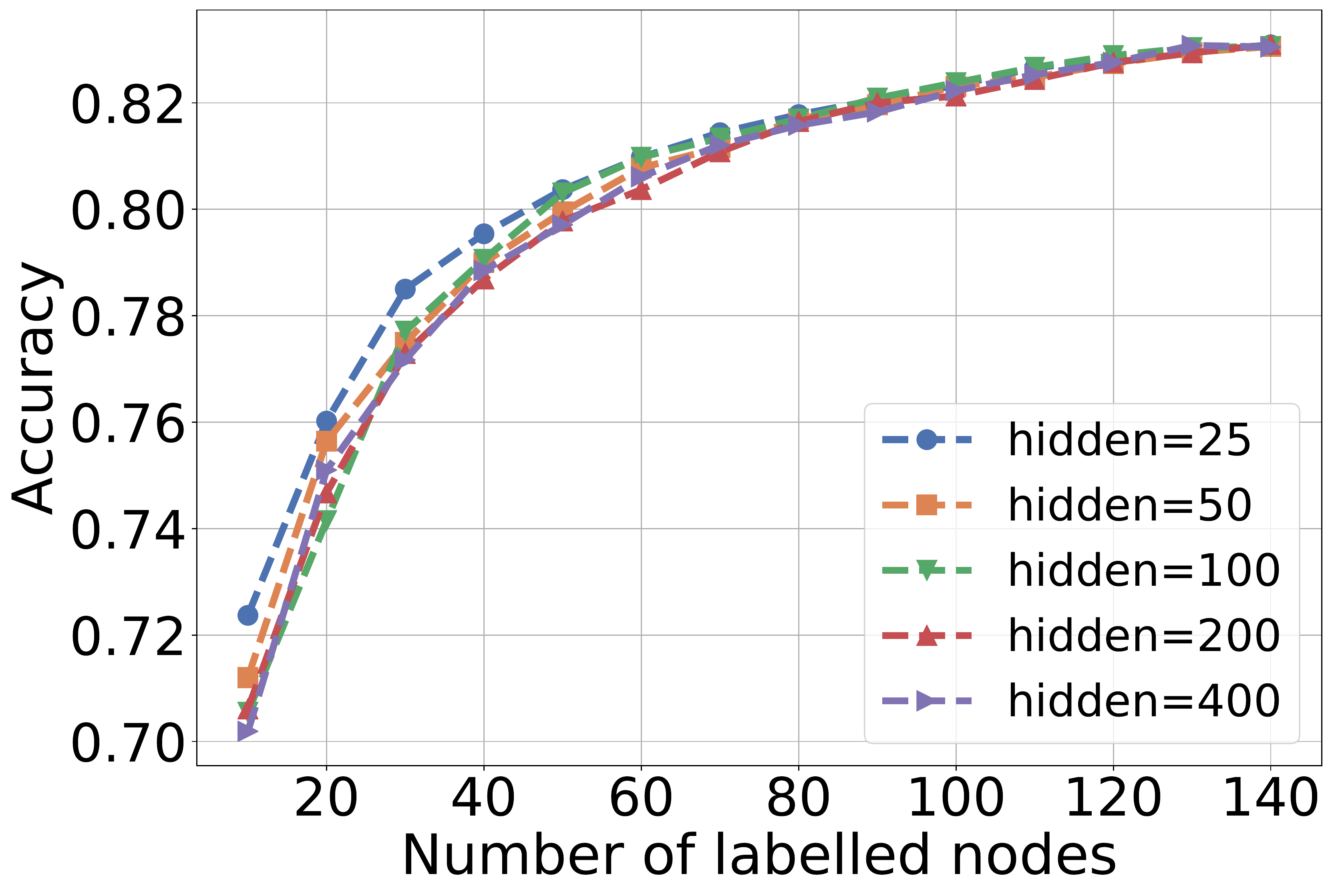}
    \caption{Cora}
  \end{subfigure}
  \begin{subfigure}[b]{0.45\columnwidth}
    \centering
    \includegraphics[width=\linewidth]{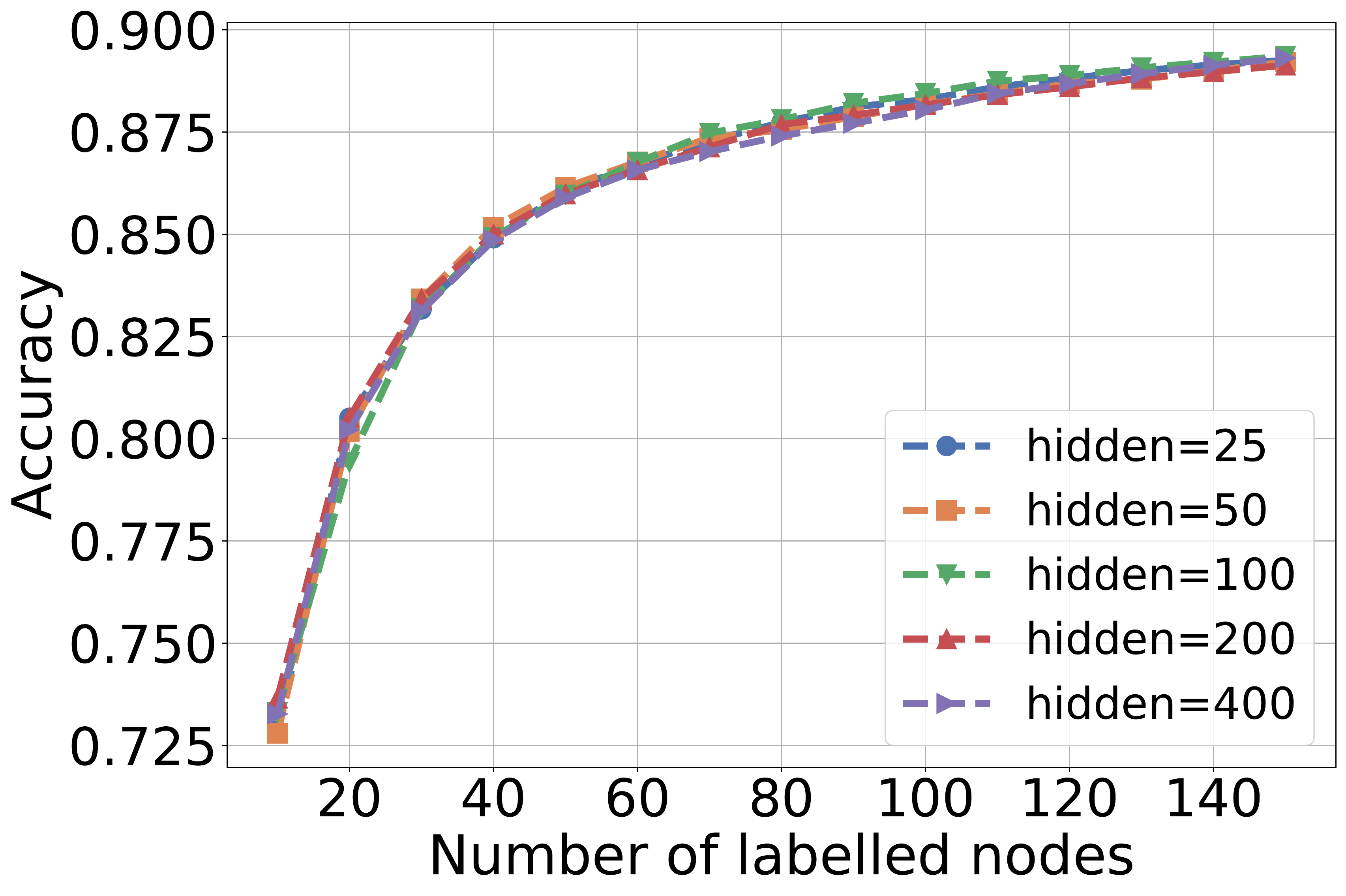}
    \caption{Co-CS}
  \end{subfigure}
  \caption{The Micro-F1 Scores on Cora and Co-CS datasets with different number of hidden units $l'$.}
  \label{fig:hyper_hidden}
\end{figure}

\begin{figure}
    \centering
  \begin{subfigure}[b]{0.45\columnwidth}
    \centering
    \includegraphics[width=\linewidth]{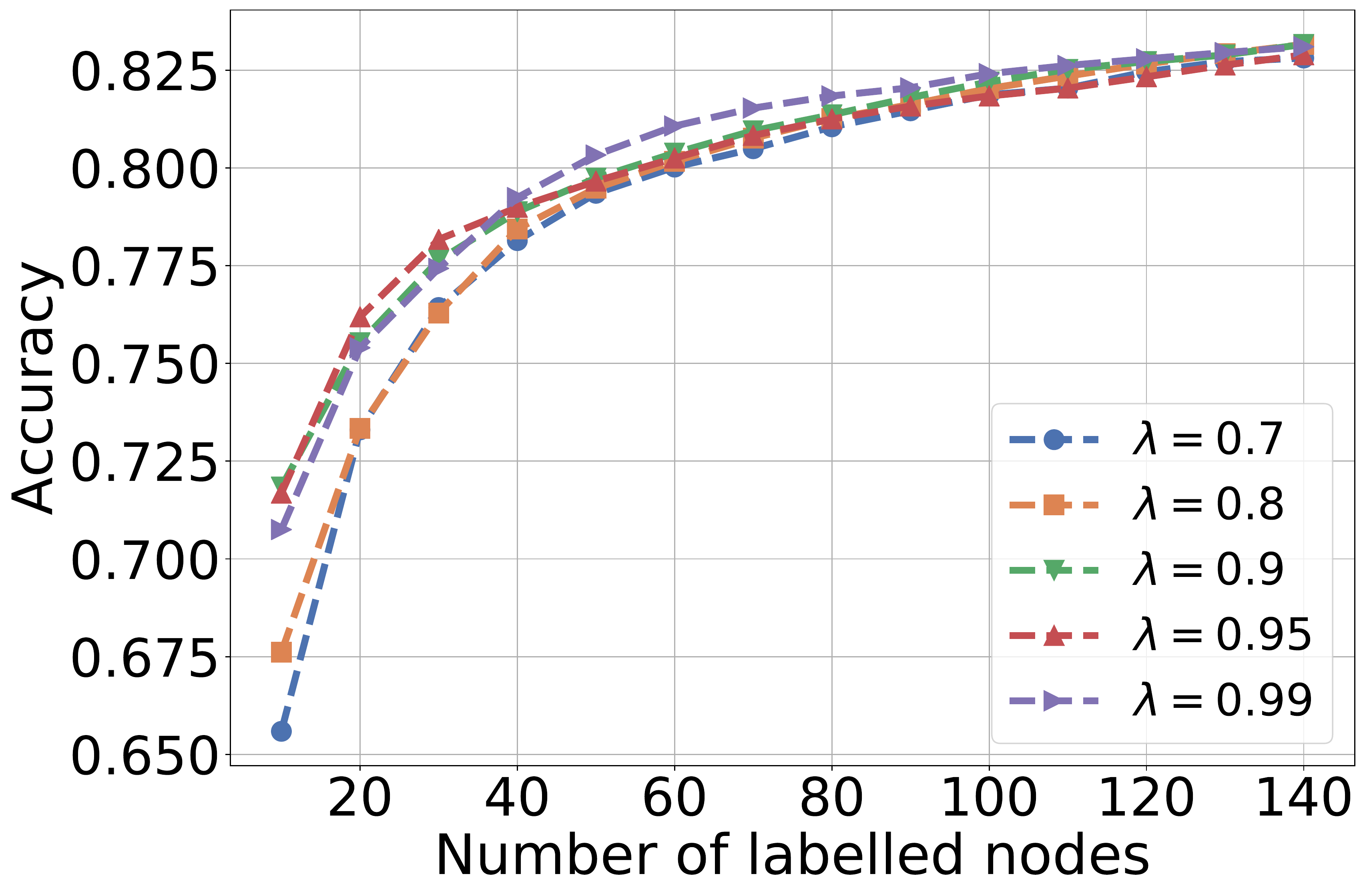}
    \caption{Cora}
  \end{subfigure}
  \begin{subfigure}[b]{0.45\columnwidth}
    \centering
    \includegraphics[width=\linewidth]{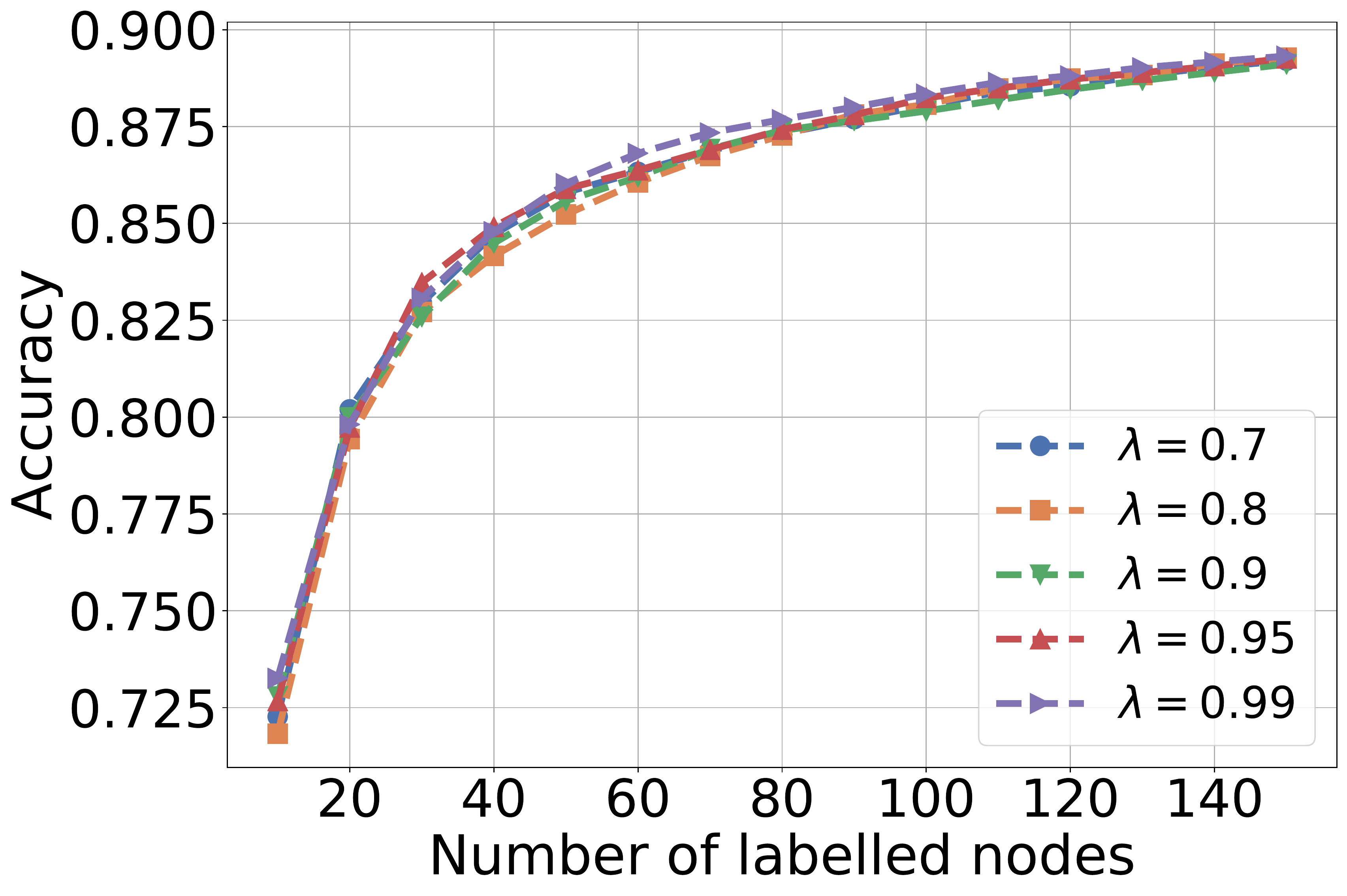}
    \caption{Co-CS}
  \end{subfigure}
  \caption{The Micro-F1 Scores on Cora and Co-CS datasets with different $\lambda$.}
  \label{fig:hyper_lambda}
\end{figure}
We study the effect of varying the parameters in our method, including the number of hidden units $l'$ in our classifier and $\lambda$ for controlling the dynamic feature combination (Equation \ref{eq: lambda}). 
The results are demonstrated in Figure \ref{fig:hyper_hidden} and \ref{fig:hyper_lambda}. 
We alter the number of hidden units $l'$ among $\{25, 50, 100, 200, 400\}$ and $\lambda$ among $\{0.7, 0.8, 0.9, 0.95, 0.99\}$. 
As we can see in Figure \ref{fig:hyper_hidden}, changing $l'$ has no much difference in terms of performance on Co-CS, while $l'=25$ performs best on Cora when the number of labelled nodes is less than 60. 
However, when it is larger than 60, different $l'$ generate similar performances on Cora. These results show that our method is not sensitive to the number of hidden units $l'$. 

For the parameter $\lambda$, increasing $\lambda$ from $0.7$ to $0.99$ generally increases the Micro-F1 score on Cora, especially when the number of labelled nodes is small. In the meantime, $\lambda=0.99$ still provides slight better performance than $\lambda=0.7$ on Co-CS. To conclude, we should avoid relatively low $\lambda$, e.g., lower than 0.8, to prevent inferior performance when the labelling budget is small.

\subsection{Experiment Environments}
We conduct all the experiments on a hardware platform with Intel(R) Xeon(R) Gold 6240 CPU @ 2.60GHz and a single GeForce RTX 2080 Ti GPU. We use the following software packages used for implementing LSCALE and other baselines: CUDA 10.1, Python 3.7.6, PyTorch 1.4.0, TensorFlow 1.15.3, NumPy 1.18.1, SciPy 1.4.1.





\end{document}